\documentclass{article}
\PassOptionsToPackage{numbers}{natbib}

\usepackage[preprint]{neurips_2023}

\usepackage[utf8]{inputenc} 
\usepackage[T1]{fontenc}    
\usepackage{hyperref}       
\usepackage{url}            
\usepackage{booktabs}       
\usepackage{amsfonts}       
\usepackage{nicefrac}       
\usepackage{microtype}      
\usepackage{xcolor}         

\usepackage[pdftex]{graphicx} 
\usepackage{caption}
\usepackage{subcaption}

\let\oldsim\sim 
\renewcommand{\sim}{{\oldsim}}

\title{Exploring Learned Representations of Neural Networks with Principal Component Analysis}

\author{%
Amit Harlev \\
Center for Applied Mathematics \\
Cornell University \\
Ithaca, NY 14853 \\
\texttt{ah843@cornell.edu} \\
\And
Andrew Engel \\
Pacific Northwest National Laboratory \\
Richland, WA 99354 \\
\texttt{andrew.engel@pnnl.gov}
\And
Panos Stinis\\
Pacific Northwest National Laboratory \\
Richland, WA 99354 \\
\texttt{panagiotis.stinis@pnnl.gov}
\And
Tony Chiang \\
Pacific Northwest National Laboratory \\
University of Washington \\
Seattle, WA 98195\\
\texttt{tony.chiang@pnnl.gov}
}

\begin{document}

\maketitle

\begin{abstract}
    Understanding feature representation for deep neural networks (DNNs) remains an open question within the general field of explainable AI. We use principal component analysis (PCA) to study the performance of a k-nearest neighbors classifier (k-NN), nearest class-centers classifier (NCC), and support vector machines on the learned layer-wise representations of a ResNet-18 trained on CIFAR-10. We show that in certain layers, as little as $20\%$ of the intermediate feature-space variance is necessary for high-accuracy classification and that across all layers, the first $\sim 100$ PCs completely determine the performance of the k-NN and NCC classifiers. We relate our findings to neural collapse and provide partial evidence for the related phenomenon of intermediate neural collapse. Our preliminary work provides three distinct yet interpretible surrogate models for feature representation with an affine linear model the best performing. We also show that leveraging several surrogate models affords us a clever method to estimate where neural collapse may initially occur within the DNN.
\end{abstract}

\section{Introduction}

In the past several years, DNNs have become a common tool in many scientific fields and real-world applications. As their use becomes more widespread, it is more important now than ever to better our understanding of these models. One way this can be accomplished is by studying their learned representations. This topic has been explored by many papers in recent years, including methods such as linear probing (\cite{alain2016understanding, cohen2018dnn, raghu2017svcca, montavon2011kernel}), studying the dimensionality of the manifold underlying the activations (\cite{ansuini2019intrinsic, recanatesi2019dimensionality, zhang2017local}), and studying the geometry of the learned representations (\cite{papyan2020prevalence}).
 
In this paper, we return to a classical tool for data analysis, \textit{principal component analysis}, to help us better understand the learned representations present in DNNs. While several papers have used PCA to study learned representations (e.g. \cite{montavon2011kernel, raghu2017svcca}), we are the first to study in depth the performance of multiple surrogate models using varying number of PCs across an entire CNN. We train a k-nearest neighbors classifier (k-NN), a nearest class-center classifier (NCC), and a support vector machine (SVM) on each residual block's activations after projecting down to the first $d$ principal components (PCs) and make qualitative observations based on the results. Studying a pretrained ResNet-18 on the CIFAR10 dataset, we observed that:

\begin{enumerate}
    \item The SVM matches or outperforms the k-NN and NCC across the network.
    \item The best possible performance of k-NN and NCC models on intermediate layer activations are completely determined by the first $\sim 100$ PCs. In fact, the k-NN model seems to overfit as additional PCs are used.

    \item The low-variance PCs of intermediate layers contain meaningful information that improves SVM performance.

    \item In the latter half of the network, the PCs necessary for $90\%$ of the classification accuracy account for only $20\%$-$40\%$ of the variance.
\end{enumerate}

\section{Related work}
\label{headings}
\begin{figure}
    \centering
    \includegraphics[height=0.8\linewidth, angle=90]{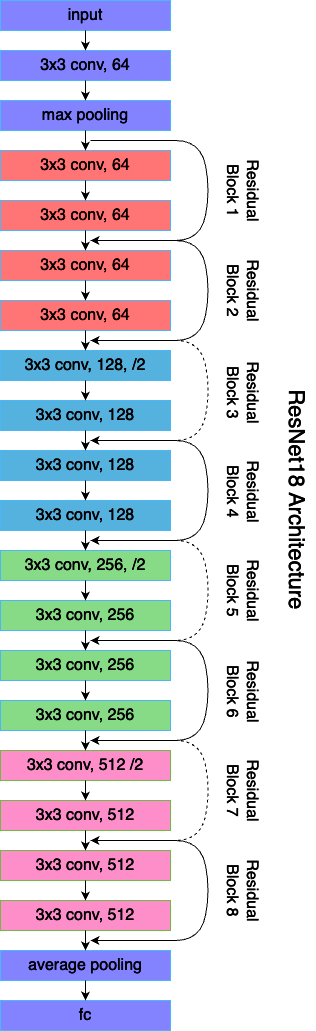}
    \captionsetup{belowskip=-16pt}
    \caption{Diagram showing ResNet-18 architecture with residual blocks labeled.}
    \label{fig:architecture}    
\end{figure}
\paragraph{Probing intermediate layers.} The idea behind classifier probes is that we can learn more about the behavior of intermediate layers, and thus neural networks in general, by studying the suitability of the intermediate representations for the desired task. The term "probe" was introduced by \cite{alain2016understanding}, who observed that the measurements of linear probes monotonically and gradually increased on trained networks the deeper they were in the network. \cite{cohen2018dnn} observed that k-NN, SVM, and logistic regression probes all match the performance of a DNN in the last layer and that the k-NN predictions are almost identical to those of the DNN. {\cite{montavon2011kernel} projected each layer's activations down to the first $d$ (RBF) kernel principal components before training linear classifiers. They studied changes in performance as architecture, hyperparameters, and $d$ were varied. While {\cite{montavon2011kernel}} studied early CNN architectures, we study behaviors of modern residual networks. {\cite{raghu2017svcca}} introduced SVCCA, a technique combining SVD and canonical correlation analysis, to study the relationships between representations coming from different layers, training epochs, and architectures. They show that ``trained networks perform equally well with a number of directions just a fraction of the number of neurons with no additional training, provided they are carefully chosen with SVCCA.''}

\paragraph{Intrinsic dimension (ID) of neural network representations.} Another approach to understanding the learned representations of DNNs has been to study their dimensionality across the network. \cite{zhang2017local} used tangent plane approximations to estimate the dimension of feature maps and observed that they declined quickly across the network. More recently, \cite{ansuini2019intrinsic} and \cite{recanatesi2019dimensionality} estimated IDs several orders of magnitude smaller than those of \cite{zhang2017local} using non-linear methods designed for curved manifolds. They also observed the layerwise ID profile to have a ``hunchback'' shape where the ID first increases and then drastically decreases. \cite{ansuini2019intrinsic} compared against ``PC-ID'', the number of PCs required to explain $90\%$ of the variance in the activations. They observed that (1) layerwise PC-ID profiles were qualitatively the same in trained and untrained networks and (2) the PC-IDs were one to two orders of magnitude greater than IDs estimated with non-linear methods. Using this, they argued that the activations must lie on a highly curved manifold. While this may be the case, we show that PCA can in fact help find interesting structures in learned representations. Additionally, we show that while the underlying manifold may be highly curved, it exists in a low-dimensional subspace that can be found using PCA.

\paragraph{Neural collapse.} First defined by \cite{papyan2020prevalence}, neural collapse is a phenomenon observed in the last layer activations of deep neural networks characterized by several properties, two of which are: \textbf{(NC1)} within-class variability collapses to zero and the activations collapse towards their class means and \textbf{(NC4)} the DNN classifies each activation using the NCC decision rule. Since then, there has been significant interest in investigating this phenomenon, including several papers exploring whether this phenomenon manifests in earlier layer's activations (\cite{rangamani2023feature, galanti2022implicit, ben2022nearest}).
Both \cite{galanti2022implicit} and \cite{ben2022nearest} study the performance of the NCC classifier across the layers of a neural network and observe an increase in performance the deeper the layer is in the network and the more training epochs used. \cite{rangamani2023feature} shows that the within-class covariance decreases relative to the between-class covariance as you move deeper into a trained DNN. 
\begin{figure}
  \centering
  \hfill
  \begin{subfigure}[b]{0.325\linewidth}
         \centering
         \includegraphics[width=\textwidth]{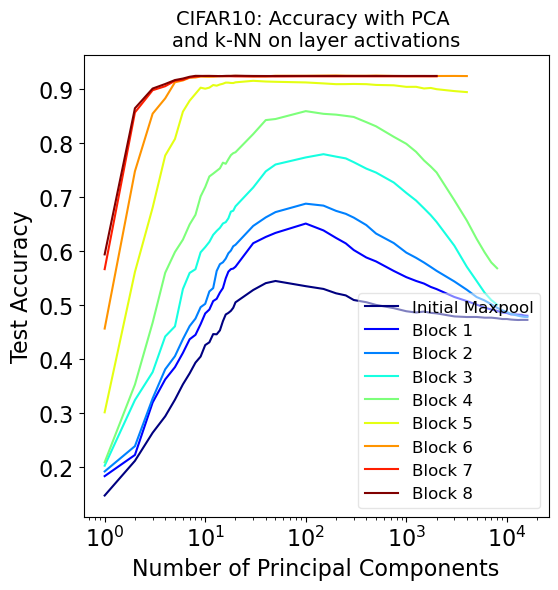}
         \caption{}
         \label{fig:KNNAccByDim}
 \end{subfigure} %
 \begin{subfigure}[b]{0.325\linewidth}
         \centering
         \includegraphics[width=\textwidth]{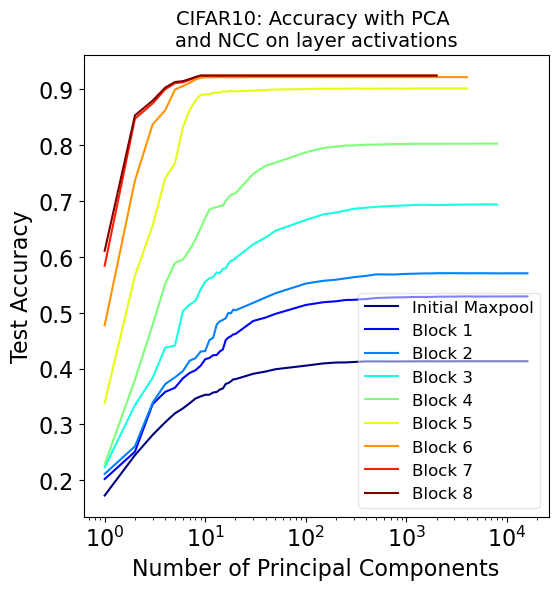}
         \caption{}
         \label{fig:NCCAccByDim}
 \end{subfigure} %
 \begin{subfigure}[b]{0.325\linewidth}
         \centering
         \includegraphics[width=\textwidth]{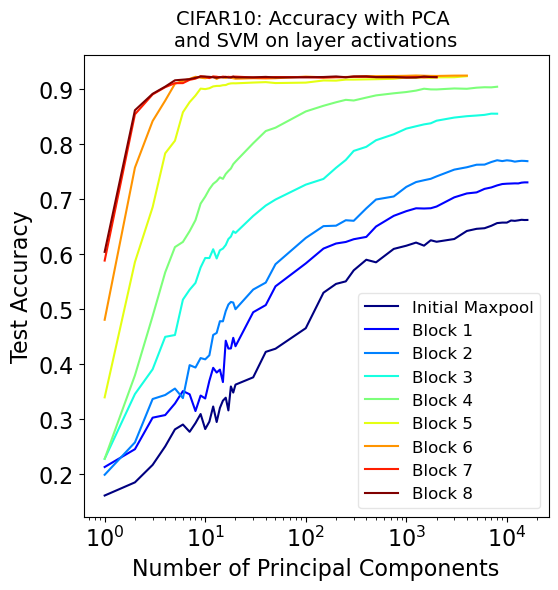}
         \caption{}
         \label{fig:SVMAccByDim}
 \end{subfigure}
 \hfill
 \captionsetup{belowskip=-10pt}
 \caption{Performance of 10-NN (\protect\subref{fig:KNNAccByDim}), NCC (\protect\subref{fig:NCCAccByDim}), and SVM (\protect\subref{fig:SVMAccByDim}) after projecting activations from each residual block onto first $d$ principal components.}
 \label{fig:AccuracyByDim}
\end{figure}
\section{Experiment}
We used a pre-trained (\cite{huy_phan_2021_4431043}) ResNet-18 (\cite{he2016deep}) with a test accuracy of $92.5\%$ on the CIFAR-10 dataset (\cite{krizhevsky2009learning}). For a given layer, we standardized (mean zero, std one) the activations from the training data and then used PCA to project onto the first $d$ PCs. We trained a 10-nearest neighbors model, nearest class-center model, and soft-margin support vector machine on the resulting data and then used them to classify the test data after applying the same standardization and projection learned from the training data. This was done for each \(d = 1-20\), \(30\), \(40\), \(50\), \(100\), \(150\), \(200\), \(250\), \(300\), \(400\), \(500\), \(750\), \(1000\), \(1250\), \(1500\), \(1750\), \(2000\) and subsequently at intervals of $1000$ until reaching the size of the layer. \autoref{fig:AccuracyByDim} shows the accuracy by number of PCs for each model. For each model and layer, we also found the minimum number of PCs needed to attain at least $90\%$ of the best accuracy attained at that layer and by that model, as well as the variance explained by those PCs. For example, if model X's highest attained accuracy on layer Y was $96\%$, we found the minimum number of PCs for which model X attained $96\%*0.9 = 86.4\%$ accuracy. This is shown in {\autoref{fig:for90Accuracy}}.
We considered the activations output by the initial max pooling layer and each of the eight residual blocks present in a ResNet-18--- see \autoref{fig:architecture}.

\section{Results}
Looking at \autoref{fig:AccuracyByDim}, we see that up until block 4, each of our three models exhibits different behaviors as we increase the number of PCs, and that from block 5 onwards, all three models exhibit qualitatively identical behavior. Up until block 4, the k-NN model's (\autoref{fig:KNNAccByDim}) accuracy increases up to $\sim 100$ PCs before decreasing significantly, a sign that it may be overfitting. On the other hand, the NCC model (\autoref{fig:NCCAccByDim}) achieves maximum accuracy at around the same point, but then remains unchanged as more PCs are used. The SVM (\autoref{fig:SVMAccByDim}) performs similarly to the k-NN for the first $\sim 100$ PCs, but continues to improve in accuracy as the number of PCs increases. It also achieves the best performance with the original activations (i.e. before projection) across all layers. All three models see steady increases in accuracy as we move deeper into the network. On blocks 5 onwards, all three models see a sharp, almost identical spike up to the true accuracy of the DNN between one and ten PCs, followed by no change in accuracy beyond that. 

\begin{figure}
  \centering
  \hfill
  \begin{subfigure}[b]{0.44\linewidth}
         \centering
         \includegraphics[width=\textwidth]{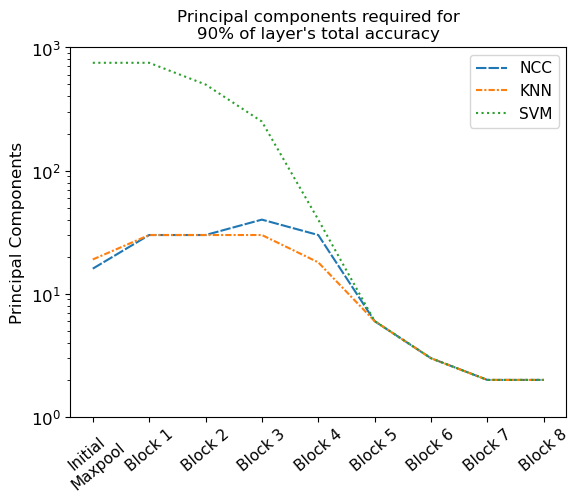}
         \caption{}
         \label{fig:PCFor90Acc}
 \end{subfigure}
 \hfill
 \begin{subfigure}[b]{0.44\linewidth}
         \centering
         \includegraphics[width=\linewidth]{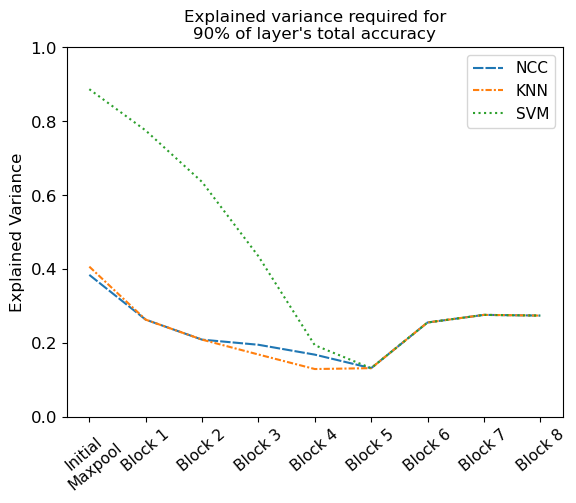}
         \caption{}
         \label{fig:ExplainedVarianceForAcc}
 \end{subfigure} %
 \hfill
  \captionsetup{belowskip=-15pt}
 \caption{For each model: number of PCs (\protect\subref{fig:PCFor90Acc}) and the percentage of variance explained by those PCs (\subref{fig:ExplainedVarianceForAcc}) needed to attain $90\%$ of maximum classification accuracy at each residual block.}
 \label{fig:for90Accuracy}
\end{figure}

In \autoref{fig:PCFor90Acc} we see a ``hunchback'' profile for the NCC model (and to a lesser degree, the k-NN model) that matches the ``hunchback'' ID profile that {\cite{ansuini2019intrinsic}} observed using a non-linear dimensionality estimator. On the other hand, the SVM, the only affine-linear method we studied, exhibits a completely different profile starting very high and then monotonically decreasing. We observe that, just as in \autoref{fig:AccuracyByDim}, all three models exhibit identical profiles for blocks 5 through 8 and that, excluding block 5, they require \textit{only $2$-$3$ PCs to attain $90\%$ of the accuracy of the DNN}. \autoref{fig:ExplainedVarianceForAcc} shows us that in the latter half of the network, only $20\%$ to $40\%$ of the variance is needed for accurate classification, and that this holds true across the entire network for the non-linear models.
\section{Discussion and conclusion}

While the performance of the k-NN and NCC models is determined by the first $\sim 100$ PCs, the SVM's performance increases with the number of PCs up to using the whole space. When considered along with the observations of intermediate neural collapse of \cite{rangamani2023feature}, this could perhaps point to there being a ``partially collapsed'' subspace in each layer that determines the behavior of the k-NN and NCC models, while the SVM also accounts for information helpful to classification in the low variance subspaces. In particular, this means that the low-variance subspaces contain meaningful information and not just noise. Additionally, it is interesting to note that the SVM, an affine-linear model, is the most robust and best performing across all learned representations of the DNN. While all three models contribute to our intuitive understanding of how the representation is changing across the network, the SVM's accuracy suggests that applications using learned representations might benefit most from simpler models.

The behavior in blocks 5-8 can also be explained by neural collapse. That is, the network reaches a ``fully collapsed state'' at block 5 in which all activations are approximately equal to their class means, so all three classifiers perform equally well on very few PCs. Note that had we only trained one surrogate model, it would not be clear between which layers the network was ``fully collapsing''. However, with three models, \autoref{fig:AccuracyByDim} and \autoref{fig:for90Accuracy} clearly show that this collapse occurs between the fourth and fifth residual blocks. Identifying this ``collapsing'' layer could be a useful tool for understanding mis-classified training data, as most of the information used by the DNN for classification is only present prior to that layer. The notion of intermediate neural collapse is further supported by the fact that the number of PCs needed for good classification with the SVM decreases monotonically across the network and that the variance necessary for accurate classification (by all models) decreases until block 5, which is where we see ``full collapse''.

Since k-NN, NCC, and PCA are all very well understood, the fact that these non-linear models display the same profile in  \autoref{fig:PCFor90Acc} as observed by {\cite{ansuini2019intrinsic}} provides us a more interpretable way to think about this ``hunchbacked'' behavior. Additionally, since the non-linear methods required only $\sim 100$ PCs or less throughout the network, this implies that the curved manifold underlying the activations most likely lives within a relatively low-dimensional subspace, which can be found using PCA.

Lastly, while it is common to select the number of PCs to keep using metrics such as accounting for $90\%$ of variance—as seen in \cite{ansuini2019intrinsic} and \cite{raghu2017svcca}—\autoref{fig:ExplainedVarianceForAcc} shows that this may not be the best approach for analyzing learned representations, as the majority of the variance is not necessary for classification.

In this paper, we study learned representations of a ResNet-18 using PCA and observe multiple interesting behaviors. We hope that our work provides new intuition and inspires more experiments into the behavior and structure of learned representations, as well as demonstrates that there may still be more for us to learn about these complex models using simple techniques.

\section{Acknowledgements}
AH, AE, and TC were partially supported by the Mathematics for Artificial Reasoning in Science (MARS) initiative via the Laboratory Directed Research and Development (LDRD) Program at PNNL. PS was partially supported from the U.S. Department of Energy, Advanced Scientific Computing Research program, under the Scalable, Efficient and Accelerated Causal Reasoning Operators, Graphs and Spikes for Earth and Embedded Systems (SEA-CROGS) project (Project No. 80278). PNNL is a multi-program national laboratory operated for the U.S. Department of Energy (DOE) by Battelle Memorial Institute under Contract No. DE-AC05-76RL0-1830

\nocite{*}
\bibliographystyle{abbrv}
\bibliography{refs}
\end{document}